\documentclass[10pt, conference, compsocconf]{IEEEtran}
\usepackage{graphicx}
\usepackage[cmex10]{amsmath}
\usepackage{amssymb}
\usepackage{cite}
\usepackage{bm}
\usepackage{caption}
\usepackage{subcaption}
\usepackage{fixltx2e}

\newcommand{\tab}{\hspace*{2em}}
\newcommand{\PP}{\mathbb{P}} 
\newcommand{\EE}{\mathbb{E}}

\begin{document}
\title{Infinite Mixed Membership Matrix Factorization}
\author{\IEEEauthorblockN{Avneesh Saluja\IEEEauthorrefmark{1},
Mahdi Pakdaman\IEEEauthorrefmark{2},
Dongzhen Piao\IEEEauthorrefmark{1} and
Ankur P. Parikh\IEEEauthorrefmark{3}}
\IEEEauthorblockA{\IEEEauthorrefmark{1}Electrical \& Computer
  Engineering Dept.\\
 Carnegie Mellon University, Pittsburgh, PA}
\IEEEauthorblockA{\IEEEauthorrefmark{2}Intelligent Systems Program\\
University of Pittsburgh, Pittsburgh, PA}
\IEEEauthorblockA{\IEEEauthorrefmark{3}Machine Learning Dept.\\
Carnegie Mellon University, Pittsburgh, PA}
}

\maketitle
\begin{abstract}

Rating and recommendation systems have become a popular application area
for applying a suite of machine learning techniques.  Current
approaches rely primarily on probabilistic interpretations and
extensions of matrix factorization, which
factorizes a user-item ratings matrix into latent user and item
vectors.  Most of these methods fail to model
significant variations in item ratings from otherwise similar users,
a phenomenon known as the ``Napoleon Dynamite'' effect.  Recent
efforts have addressed this problem by adding a contextual bias term
to the rating, which captures the mood under which a user
rates an item or the context in which an item is rated by a user.  In this work, we extend this model in a
nonparametric sense by learning the optimal number of moods or
contexts from the data, and derive Gibbs sampling inference procedures
for our model.  We evaluate our approach on the
MovieLens 1M dataset, and show significant improvements over the optimal
parametric baseline, more than twice the improvements previously
encountered for this task.  We also extract and evaluate a DBLP
dataset, wherein we predict the number of papers co-authored by two
authors, and present improvements over the parametric baseline on this
alternative domain as well.  
\end{abstract}

\section{Introduction}

Generating recommendations for users based on their previously
revealed preferences and the ratings of similar users has increased
in popularity as a machine learning problem, primarily due to the activities of internet
companies such as Amazon, or the coverage of competitions such as the
Netflix prize.  Most of these methods have at their core a matrix
factorization-based approach.  We can view the users and their ratings
for items as a $U \times M$ matrix {\bf R} (subsequently referred to as the ratings matrix or
the user-item matrix), which contains the ratings for $U$ users over
$M$ items, and where every item is rated by at least one user.  The idea
is to factorize {\bf R} into its latent factors: ${\bf R \approx A^TB}$, where
${\bf A} \in \mathbb{R}^{D \times U}$ is the latent user matrix, and
${\bf B} \in\mathbb{R}^{D \times M}$ is the latent item matrix.  These lower
dimensional representations reflect the intuition that ratings are
based on a small number of factors (with dimension $D$), and the idea
is to complete the matrix using this low-dimensional formulation, by
taking the inner product between user and item latent factors for
items not rated by users during the prediction or ``matrix
completion'' step.   

This general approach is also known as collaborative filtering, since
one not only utilizes the user's previous preferences but also the
preferences of similar users (hence the ``collaborative'' aspect).
Probabilistic formulations of matrix factorization
for recommendation systems generally improve upon the performance of
their non-probabilistic counterparts, by treating the latent factors
as random variables.  However, such methods are
unable to model context explicitly, since most probabilistic models of
this form generate the ratings from a normal distribution with the
mean set to the
inner product between the user and item latent factors.
As a result, phenomena like the \emph{Napoleon Dynamite effect},
originally proposed in the context of the Netflix prize, wherein
certain movies (or more generally, items) receive high variance
ratings from users that otherwise rate other items similarly (the
movie \emph{Napoleon Dynamite} being a prime example), cannot be
handled well in the traditional probabilistic matrix factorization
framework.  

The recently proposed mixed membership matrix factorization (M$^3$F)
model \cite{Mackey2010} aims to tackle this problem by explicitly modeling
ratings not only with static latent factors, but with an additional
bias term that takes into account the context of the rating.  M$^3$F
represents this bias term by modeling it through a type of mixed membership
stochastic blockmodel: each user and each item can be represented as a
discrete distribution over user and item topics, representing contexts
or moods under which users rate items.  When a user $u$ rates an item
$j$, the contextual bias of this rating is a function of the user's
bias for the item's topic for that particular rating (intuitively, the
context under which item $j$ is being rated).  It is also a function of the item's
bias for the user's topic for that rating (intuitively, the mood of the
user $u$ at the time of rating).  Thus, M$^3$F models a user's
rating for an item as a function of both the static latent factors and
the user-item group and user group-item based contextual biases. 

The M$^3$F model however, is limited in the sense that the number of user
and item topics for the contextual bias term is set beforehand and not
learned from the data.  In our proposed extension to this model, which
we refer to as the \emph{infinite mixed membership matrix
  factorization} model (iM$^3$F), we sample the user and item topic
assignments from an infinite-dimensional prior, allowing us to learn
the optimal number of user and item topics from the data.  Ideally, we
would like the user and item topics to be shared across all
users and items, but that each user and item maintain their own topic proportions
over these topics.  Thus, we propose to use a hierarchical Dirichlet
process prior on the number of item and user topics, and
sample assignments through a Chinese restaurant franchise
representation. A Gibbs sampling procedure is derived and implemented for this
model, and we evaluate its performance on two datasets from different
domains: a movie ratings dataset (MovieLens), and a publication
co-authorship database that we extracted ourselves from the DBLP
database.  Compared to the baseline M$^3$F model on the former
dataset, we decrease root mean square error (RMSE) by 0.0065, which is
two times higher than M$^3$F's improvement over its baseline.  We
also significantly outperform M$^3$F on the DBLP dataset.  
\section{Related Work}
As a way to attack recommender system problems, matrix factorization
has been one of the popular approaches along with content-based
filtering \cite{Herlocker2004}, and in recent years has proven to be
the superior alternative of the two \cite{Koren2009}.  Probabilistic
matrix factorization (PMF) was first introduced by Salakhutdinov and Mnih
\cite{Salakhutdinov08a}, whereby ratings are generated from a Gaussian
with a mean computed as the dot product of the user and item latent
factors, with fixed variance.  A fully Bayesian version of PMF (BPMF) was
subsequently proposed by the same authors \cite{Salakhutdinov08b},
where they assumed the latent factors were themselves generated by
Gaussian distributions, with normal-Wishart priors on the
hyperparameters.  

Our work is primarily based on extending the M$^3$F model
\cite{Mackey2010} .  In M$^3$F, the user and item latent factors are
modeled as in BPMF, but when generating a rating, in addition to using
the dot product of the latent factors as the mean, we add a contextual
bias term.  More formally, we generate a rating by a user $u$ for an
item $j$ as\footnote{note that this is one of two variants
    proposed in the original paper, the Topic-Indexed Bias (TIB)
    model.  The other model, the Topic-Indexed Factor (TIF) model, was
    found to underperform the TIB model.}:
\begin{align}
  r_{uj} \sim \mathcal{N} (\chi_0 + c_u^k + d_j^i + {\bf a}_u \cdot
  {\bf b}_j, \sigma^2)
\end{align}
where $c_u^k$ is the latent rating bias of user $u$ under item topic
$k$, $d_j^i$ denotes the bias for item $j$ under user topic $i$,
$\chi_0$ is a fixed global bias, and ${\bf a}_u$ and ${\bf b}_j$ are
the latent user and item factors respectively.  The M$^3$F framework's concept of modeling user and item biases through user-item topic
and item-user topic interactions is derived from mixed membership
stochastic blockmodels (MMSBs) \cite{Airoldi2008}, which itself is influenced by the original
mixed membership topic modeling framework \cite{Blei2003}, and stochastic
block models \cite{Kemp2006,Nowicki2001}.   

Collaborative topic regression \cite{Wang2011} also combines matrix factorization and mixed
membership topic modeling, but their method solves a different problem, the cold start issue of
out-of-matrix predictions, rather than contextual bias.  

There has been a good amount of work on general non-parametric
techniques in matrix factorization and in particular, applications in
collaborative filtering.  However, all of these approaches focus on
the parameters corresponding to the number of latent factors, and aim to learn the optimal
latent dimension from the data.  \cite{Yu2009} do not use a
probabilistic approach and instead use SVD, whereas \cite{Ding2010,
  Xu2012} adopt an Indian Buffett Process to model the latent factor
dimensionality, and implement variational inference-based solutions.
We note that such non-parametric approaches are orthogonal to our
method, since they target the latent dimensionality, and can thus be
easily integrated and combined with our approach.  
\section{Approach}
Our proposed method, iM$^3$F, is based on extending the M$^3$F approach for probabilistic modeling of
recommendation systems.  In iM$^3$F, we add an infinite-dimensional
prior for user and item topics.  This extension is realized in
our model by using a nonparametric Bayesian prior for the topic distributions, providing the model the
ability to have a potentially infinite number of user and item topics. We present the schematic and symbolic views of our model, and
derive a Gibbs sampling inference scheme using the Chinese
restaurant representation of the nonparametric prior.  

\subsection{Chinese Restaurants}
\label{sec:mainapproach}
The nonparametric prior on the
mixtures over topics for users and items is motivated by its ability
to allow the data to dictate the optimal number of user and
item groups, instead of parameterizing these values beforehand.  Using
such a prior makes it unnecessary to conduct extensive ``tuning''
experiments for the number of topics, and is arguably conceptually simpler and
more elegant.  We provide a
brief background of the ideas, with an emphasis on the intuition.  For
a more rigorous introduction, we refer the reader to \cite{Ishwaran2002}.  For the
purposes of exposition, we discuss the case of users being defined as
mixtures over user topics (i.e., a user is defined as a mixture of
various ``moods''), and note that the item case is completely
symmetrical.  

In the finite case, we can represent this
mixture as a multinomial distribution over the various user topics.  The conjugate
prior for a multinomial distribution is the Dirichlet distribution,
sometimes seen as a ``distribution over distributions'' or a
distribution over finitely many points on the real line.  If we
take the number of mixture components to infinity, then the conjugate
prior becomes a Dirichlet process (DP) \cite{Ferguson1973}, a distribution over an infinitely fine-grained real line.  
\begin{figure}[h]
  \begin{center}
    \includegraphics[scale= 0.35]{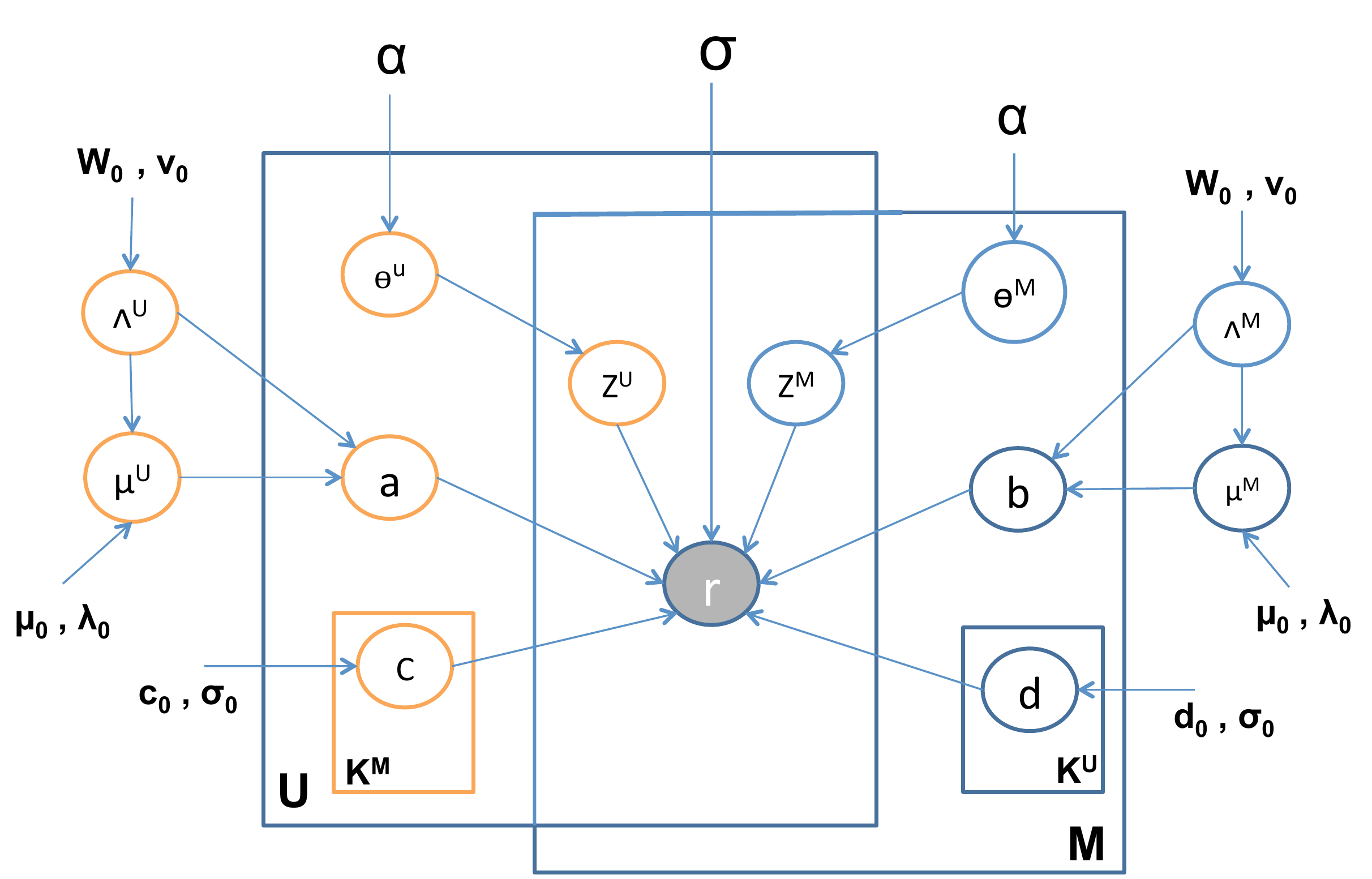}
  \end{center}
  \caption{The original M$^3$F model, as proposed in \cite{Mackey2010}.}
  \label{IMG:M3F}
\end{figure}
\begin{figure}[h]
  \begin{center}
    \includegraphics[scale= 0.35]{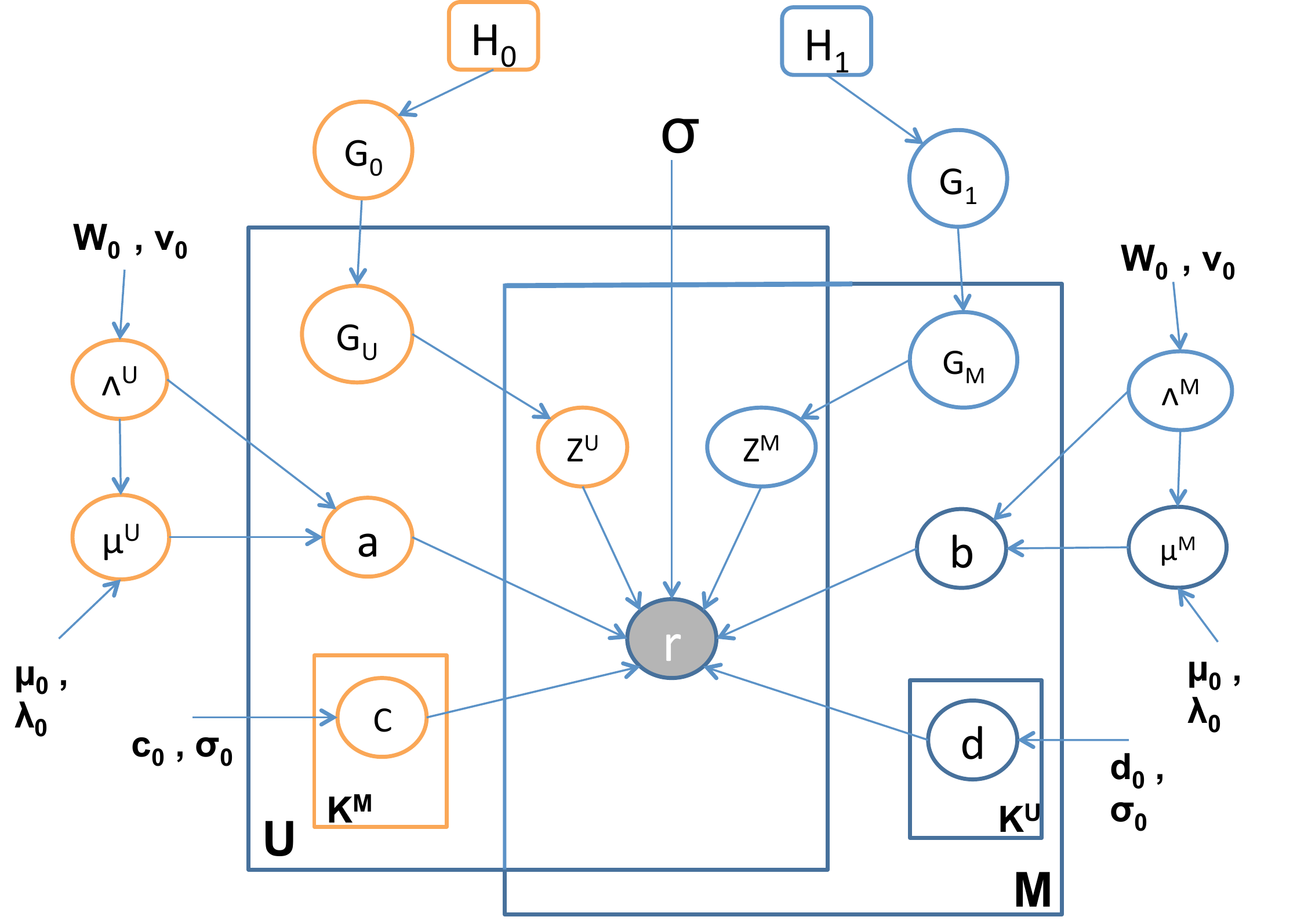} 
  \end{center}
  \caption{Our proposed iM$^3$F model with an HDP prior on the user and
  item topics.}
\label{IMG:M3F-Extended}
\end{figure}
In practice, we do not have a density defined for the
infinite-dimensional DP prior, so we define it indirectly via a sampling
scheme that is used to generate
samples from this prior \cite{Ishwaran2002}.  If we are interested in both the weights and
the partitions, then we use a stick-breaking construction to generate
samples.  Otherwise, if we are simply interested in partition or
cluster assignments for the data, we can marginalize the mixture
weights and sample the assignments directly, through a process known
as a Chinese restaurant process (CRP) \cite{Blackwell1973}.  In our situation, for every
rating $r_{uj}$, we can simply sample from the CRP prior to ascertain which
item and user topics were used to generate the contextual bias for the
rating.  

A naive way to incorporate the CRP prior is to assume that
each user and item maintain their own priors over item and
user topics\footnote{Such a ``separate'' CRP model was implemented
  and evaluated, but the results were worse than a parametric solution.}.  Ideally though, we would like each user or item
to maintain their own mixture weights over topics, but have a global
set of topics that are shared across all users and items.  This
requirement necessitates a hierarchical Dirichlet process (HDP) prior
\cite{Teh2006} on the mixture components.  In the Chinese restaurant
metaphor, we now use a Chinese restaurant franchise (CRF) prior.
Intuitively, this is a two-stage Dirichlet process, where we have one
process for a global set of ``dishes'', or topics that are shared by
all users (or all items), and another user-specific process for the
table assignments for individual customers (or ratings).  The base distribution for each user-specific CRP
is the process corresponding to the shared set of dishes, which forces
each user (or item) to maintain a distribution over the shared set of
topics.  

\subsection{Generative Process}
We present the generative process of the iM$^3$F model, depicted in
Figure \ref{IMG:M3F-Extended}.  For the purposes of comparison, the schematic view of the finite, parametric case is presented in Figure
\ref{IMG:M3F}, which corresponds to the original M$^3$F model. First, the hyperparameters
for the latent user and item factors are generated:
\begin{align*}
&\Lambda^{U} \sim \textrm{Wishart}({\bf W}_0 ,v_0) ,~\Lambda^{M} \sim
\textrm{Wishart}({\bf W}_0 ,v_0) \\
&\mu^U \sim \mathcal{N}(\mu_0 , (\lambda_0 \Lambda^U )^{-1}),~\mu^M \sim \mathcal{N}(\mu_0 , (\lambda_0 \Lambda^M )^{-1})
\end{align*}
We also generate the global probability measures that are used to
define a set of shared clusters for both the user and item topics:
\begin{align*}
G_0 \sim \textrm{DP}(\beta, H_0),~G_1 \sim \textrm{DP}(\beta, H_1)
\end{align*}
Note that we assume the DPs associated with the item and user global
dishes have the same concentration parameter $\beta$, for the purposes
of simplicity, and that the base measures are discrete and uniform.   Now that we have the hyperparameters, we can sample the latent
factor ${\bf a}_u$ for each user and the latent factor ${\bf b}_j$ for each
item by sampling from a normal distribution.  Furthermore, since we have the
global base measures, we can sample the user- and item-specific
mixture distributions from DPs with discrete base measures $G_0$ and
$G_1$:
\begin{align*}
\textrm{For each }& u\in \{1,\ldots, U\}: \\
  &{\bf a}_u \sim \mathcal{N}(\mu^U ,(\Lambda^U)^{-1}) \\
  &G_u \sim \textrm{DP}(\gamma, G_0) \\ \\
\textrm{For each }& j \in \{1, \ldots, M\}: \\
&{\bf b}_j \sim \mathcal{N}(\mu^M , (\Lambda^M)^{-1})\\
&G_j \sim  \textrm{DP}(\gamma, G_1)
\end{align*}
Once again, we assume the DPs associated with the item and
user-specific mixtures have the same concentration parameter
$\gamma$. With a slight abuse of notation, we can generate the user
group and item group topic assignment indicator variables from CRPs with appropriate base distributions:
\begin{align*}
  \textrm{For each }& \textrm{rating } r_{uj}: \\
  &z_{uj}^{U} \sim \textrm{CRP}(\gamma, G_u) \\
  &z_{uj}^{M} \sim \textrm{CRP}(\gamma, G_j) \\
\end{align*}
where $z_{uj}^{U}$ is the user cluster or topic that user
$u$ belongs to when rating item $j$ (user $u$'s mood at the time of
rating item $j$). Similarly, $z_{uj}^{M}$ is
the item cluster or topic that item $j$, when rated by user $u$,
belongs to.  Note that in practice the topic assignments are sampled
by marginalizing out the base distributions $G_u$ and $G_j$, and the
only relevant parameter is $\gamma$, but for notational clarity we
include the base measure.  Letting $k = z_{uj}^{M}$ and $i = z_{uj}^{U}$, we can generate
$c_{u}^{k}$, the bias of user $u$ for item topic $k$, and similarly
generate $d_{j}^{i}$, the bias of a specific item $j$ for the
user topic $i$ as follows:
\begin{align*}
&d_{j}^{i} \sim \mathcal{N}(\mu_i , \sigma_0^{2}) \\
&c_{u}^{k} \sim \mathcal{N}(\mu_k , \sigma_0^{2}) 
\end{align*}
Finally, we have all the pieces in place to generate the rating:
\begin{align*}
  r_{uj} \sim \mathcal{N}(\beta_{uj}^{ik}+ {\bf a}_u \cdot {\bf b}_j , \sigma^{2}) \nonumber 
\end{align*}
where $\beta_{uj}^{ik} = \chi_0 + c_u^k +  d_j^i$.  
\subsection{Inference via Gibbs Sampling}
Our sampling procedure is
similar to the M$^3$F sampling scheme, except for the cluster assignment indicators $z^U$ and $z^M$, as they
are generated from a CRF (in iM$^3$F) instead of a
multinomial with a Dirichlet prior.  

We first sample the hyperparameters given the other variables in the
model:
\begin{align}
  \Lambda^{U} | \textrm{rest}\backslash\{\mu^U\} \sim
  &\textrm{Wishart}(({\bf W}_0^{-1} + \sum_{u=1}^{U}
({\bf a}_u - \bar{\bf a})({\bf a}_u - \bar{\bf a})^T)\nonumber \\
&+\frac{\lambda_0 U}{\lambda_0 +
U}(\mu_0-\bar{\bf a})(\mu_0-\bar{\bf a})^T)^{-1}, \nu_0 + U) \label{eq:lambda_u}\\
\Lambda^{M} | \textrm{rest}\backslash\{\mu^M\} \sim
&\textrm{Wishart}(({\bf W}_0^{-1} + \sum_{j=1}^{M}
({\bf b}_j - \bar{\bf b})({\bf b}_j - \bar{\bf b})^T) \nonumber \\
&+\frac{\lambda_0 M}{\lambda_0 +
M}(\mu_0-\bar{\bf b})(\mu_0-\bar{\bf b})^T)^{-1}, \nu_0 + M) \label{eq:lambda_m}
\end{align}
where in Equation \ref{eq:lambda_u}, $\bar{\bf a} = \frac{1}{U}\sum_{u=1}^{U} {\bf a}_u$
and in Equation \ref{eq:lambda_m}, $\bar{\bf b} = \frac{1}{U}\sum_{j=1}^{M} {\bf
  b}_j$. We then sample parameters $\mu^U$ and $\mu^M$:
\begin{align}
\mu^U|\textrm{rest} \sim \mathcal{N}\left(\frac{\lambda_0 \mu_0
  +\sum_{u=1}^{U} {\bf a}_u}{\lambda_0+U},(\Lambda^U(\lambda_0+U))^{-1}\right) \nonumber \\
\mu^M |\textrm{rest} \sim  \mathcal{N}\left(\frac{\lambda_0
    \mu_0+\sum_{j=1}^{M} {\bf b}_j}{\lambda_0+M},
 (\Lambda^M(\lambda_0+M))^{-1}\right) \nonumber
\end{align}
Then, we sample the item bias for a given user $u$ and for $i \in \{1,
\dots, K^M\}$:
{\footnotesize
\begin{align}
c_u^i | \textrm{rest} \sim &\mathcal{N}\left(\frac{\tfrac{c_0}{\sigma_0^2}+ \sum_{j \in
V_u}\sigma^{-2}z_{uji}^M
\left(r_{uj} - \chi_0 - d_j^{z_{uj}^U} - {\bf a}_u \cdot {\bf b}_j\right)}{\sigma_0^{-2} + \sum_{j \in
V_u}\sigma^{-2} z_{uji}^M}, \right. \nonumber \\
&\left. \frac{1}{\sigma_0^{-2}+\sum_{j \in V_u}
    \sigma^{-2}z_{uji}^M}\right) 
\label{Eq:c} 
\end{align}}
where $V_u$ is the set of items rated by user $u$, and $z_{uji}^M$ is
an selector variable which is 1 if the item topic assignment for user
$u$ and item $j$ is $i$.  Note that $K^M$ is conceptually infinite, but in practice we go
through the item clusters that exist in the data thus far and generate
$c^i_u$.  Due to the symmetry of the user and item biases, the user
bias for an item $j$, with $i \in \{1, \dots, K^U\}$ this time ($K^U$ is also
conceptually infinite), is:
{\footnotesize
\begin{align}
d_j^i | \textrm{rest} \sim &\mathcal{N}\left(\frac{\tfrac{d_0}{\sigma_0^2}+ \sum_{u: j \in
V_u}\sigma^{-2}z_{uji}^U
\left(r_{uj} - \chi_0 - c_u^{z_{uj}^M} - {\bf a}_u \cdot {\bf b}_j\right)}{\sigma_0^{-2} + \sum_{u:j \in
V_u}\sigma^{-2} z_{uji}^U}, \right. \nonumber \\
&\left. \frac{1}{\sigma_0^{-2}+\sum_{u:j \in V_u} \sigma^{-2}z_{uji}^U}\right)
\label{Eq:d}
\end{align}}
For notational convenience, let us define the empirical means and variances of the values
sampled through Equations \ref{Eq:c} and \ref{Eq:d} as ($\mu_c$ ,
$\sigma_c$) and  ($\mu_d$ , $\sigma_d$) respectively.  

Using the conjugacy property of the Gaussian and Wishart distributions, we
can sample the latent factors. We update ${\bf a}_u$ and ${\bf b}_j$
as follows:
\begin{align}
\textrm{For each }& u, \nonumber \\
{\bf a}_u | \textrm{rest} \sim
&\mathcal{N}\left((\Lambda_u^{U*})^{-1}\left(\Lambda^U \mu^U + \right.\right.\nonumber\\
&\left.\left.\sum_{j \in V_u}
\sigma^{-2}{\bf b}_j\left(r_{uj}-\chi_0 - c_u^{z_{uj}^M} -
d_j^{z_{uj}^U}\right) \right) ,(\Lambda_u^{U*})^{-1}\right) \label{eq:a_u}\\
\textrm{For each }& j, \nonumber \\
 {\bf b}_j | \textrm{rest} \sim
 &\mathcal{N}\left((\Lambda_j^{M*})^{-1}\left(\Lambda^M \mu^M + \right.\right.\nonumber\\
&\left.\left.\sum_{u:j \in V_u}
\sigma^{-2} {\bf a}_u\left (r_{uj}-\chi_0 - c_u^{z_{uj}^M} - d_j^{z_{uj}^U}\right)\right) ,
(\Lambda_u^{U*})^{-1}\right) \label{eq:b_j}
\end{align}

where in Equation \ref{eq:a_u}, $\Lambda_u^{U*} = \Lambda^U + \sum_{j \in V_u}
\sigma^{-2}{\bf b}_j {\bf b}_j^T$, and in Equation \ref{eq:b_j}, $\Lambda_j^{M*} = \Lambda^M + \sum_{u:j \in V_u}
\sigma^{-2} {\bf a}_u {\bf a}_u^T$.  

Finally, we draw the topic assignment indicators, which is where the HDP prior
enters the picture.  In the original
M$^3$F model, user and item topic assignment indicators $z_{uj}^U$ and
$z_{uj}^M$ were sampled from a finite multinomial distribution, with
the hyperparameters of this multinomial sampled from a Dirichlet prior
multiplied by a likelihood term.  In the CRF case, user topic
and item topic assignment sampling is done via a two-stage process:
sample the global dish assignments (per table) available to the
franchise, and then sample
the local table assignments (per customer) for a restaurant.  Note
that for the problem, it is not the entire rating that is modeled by
the MMSB, but rather the \emph{rating residual}, which is the rating minus
the score predicted by the latent factors.  Therefore, let the rating
residual be defined as $x_{uj} = r_{uj} - \chi_0 - {\bf a}_u \cdot {\bf
  b}_j$.  Tying
the restaurant analogy to our problem, customers are ratings (or
rather, rating residuals, as only the $c$ and $d$ variables are
affected by the topic assignments), tables are user-specific proportions over
topics, and finally dishes are the topics themselves, in that they
represent the parameters used to generate the data (residuals).  

The corresponding Gibbs sampler thus has two sets of state variables: 
the first being the table assignment indicator $t$ of the observed
residuals to tables, and the second is the dish assignment indicator
$k$ associated with each table, which is just a group of residuals.
We sample the table assignments as follows:
\begin{align}
\PP(t_{uj}=t) &\propto N_t^{u,-j} \PP(x_{uj}|\mu_{k_t}^{-uj})
\nonumber \\
\PP(t = \bar{t}) &\propto \gamma \left(\frac{\sum_{k=1}^K 
\PP(x_{uj}|\mu_k^{-uj})M_k + \PP(x_{uj}|c_0)\beta}{\sum_{k=1}^K M_k +
\beta}\right) \nonumber
\end{align}
 where $\bar{t}$ corresponds to sampling a new table, $\beta$ and
 $\gamma$ are the hyperparameters for the CRPs corresponding to the
 dish and table assignments respectively, $N_t^{u,-j}$ is the number of
customers (rating residuals) associated with restaurant (user) $u$
sitting on existing table $t$ except the $j^{\textrm{th}}$ customer
(i.e., except the rating residual of user $u$ for item $j$), and $M_k$ is the number of
tables over all restaurants (users) on which the dish number $k$ is
being served.  In addition, $\mu_{k_t}^{-uj}$ is shorthand for the rating residual
empirical mean, computed over all residuals across all users assigned
to a table with dish $k$, except for the current residual that we are
sampling for.  $c_0$ is the prior for the bias or the residual term.
When sampling a new table for a residual $x_{uj}$, if we happen to
sample a previously nonexistent table, then we immediately sample a
new dish for that table.  

We sample dishes for each table as follows:
\begin{align}
\PP(k_{ut} = k) \propto M_k^{-ut} \prod_{t_{uj} = t} \PP(x_{uj} |
\mu_k^{-ut}) \label{eq:sample_existdish}\\
\PP(k_{ut} = \bar{k}) \propto \beta \prod_{t_{uj} = t} \PP(x_{uj} |
c_0) \label{eq:sample_newdish}
\end{align}
where $\bar{k}$ corresponds to sampling a new dish, $M_k^{-ut}$ represents the count of the tables assigned to dish
$k$ except the current table that we are sampling a dish assignment
for, and $\mu_k^{-ut}$ is shorthand for the rating residual empirical
mean, computed over all residuals across all users assigned to a table
with dish $k$ except all residuals on the current table we are
sampling for.  Note that when computing the likelihoods, we compute
the joint likelihood over all the residuals associated with the
current table, since in the dish assignment we assign a dish for a \emph{group}
of residuals.  

When computing the probability of dish assignments as per Equations \ref{eq:sample_existdish} and
\ref{eq:sample_newdish}, we used the MAP estimate in the predictive likelihood
term for our experiments (Section \ref{sec:experiments}).  To be
completely Bayesian, the actual likelihood term is:
{\footnotesize
\begin{align} 
f({\bf x}_{t_{uj} = t}|\textrm{rest}) &= \int 
\PP(\mu_k^{-ut} |\{ x_{u'j'}| k_{u'j'} = k, t_{u'j'} \neq t\}) \cdot
\nonumber \\
&\prod_{t_{uj} = t} \PP(x_{uj} | \mu_k^{-ut}) d\mu_k^{-ut} \label{eq:likelihoodintegral}\\
&= \int \PP(\theta | \mathcal{X}) \prod_{i=1}^{n} \PP(x_i | \theta)
d\theta \label{eq:likelihoodintegralsimp}\\
&\propto \int \exp \left( \frac{-(\theta - \mu_P)^2}{2 \sigma_P^2}  \right) 
\prod_{i=1}^{n} \exp \left( \frac{-(x_i - \theta)^2}{2 \sigma^2}
\right) d\theta \label{eq:likelihoodcollapse}
\end{align}  }
where in Equation \ref{eq:likelihoodintegral}, ${\bf x}_{t_{uj} = t}$ is the set of customers (residual
ratings) sitting on table $t$  (the table we are
sampling the dish for), and in Equation \ref{eq:likelihoodintegralsimp}, $n$ is the number
of customers (residuals) sitting on table $t$, $\mathcal{X} = \{ x_{u'j'}|
k_{u'j'} = k , t_{u'j'} \neq t\}$ is set of all other residuals
across all users that are associated with dish or topic $k$ (index by $l$) except 
those sitting on our special table, and $\theta = \mu_k^{-ut}$.  In
Equation \ref{eq:likelihoodcollapse}, we use the assumption that the data (residual) is
generated from a Gaussian with a conjugate prior, and $\mu_P$ and
$\sigma^2_P$ are defined as before.  From Equation \ref{eq:likelihoodcollapse},
we can say that the joint probability over $\PP(\theta, {\bf x}_{t_{uj} = t})$ is a
multivariate Gaussian, and marginalizing over $\theta$ yields a
Gaussian as well.  Using the laws of total expectation, variance, and covariance,
one can show that: 
\begin{eqnarray*}
\EE[x_i|\mathcal{X}] &=& \EE[\EE[x_i|\theta , \mathcal{X}]|\mathcal{X}] = \EE[\theta | \mathcal{X}] = \mu_P\\
\textrm{Var}[x_i|\mathcal{X}] &=& \textrm{Var}[\EE[x_i|\theta,\mathcal{X}]|\mathcal{X}] + \EE[\textrm{Var}[x_i | \theta , \mathcal{X}]|\mathcal{X}]\\
\tab &=& \sigma^2 + \sigma_P^2\\
\textrm{Cov}[x_i , x_j |\mathcal{X}] &=& \EE[\textrm{Cov}[x_i , x_j
|\theta , \mathcal{X}]|\mathcal{X}] + \nonumber \\
&&\textrm{Cov}[\EE[x_i|\theta , \mathcal{X}],
\EE[x_j|\theta , \mathcal{X}]|\mathcal{X}]\\
\tab &=& 0 + \textrm{Cov}[\theta , \theta]\\
\tab &=& \textrm{Var}[\theta] = \sigma_P^2
\end{eqnarray*}   
  
Thus, we can write the predictive likelihood term as:    
\begin{align}
f({\bf x}_{t_{uj} = t}|\textrm{rest}) \propto  
&\frac{1}{|\sum|^\frac{1}{2}} 
\exp\left( -\frac{1}{2} ({\bf x}_{t_{uj} = t}- \bm{\mu})^T \cdot
  \right.\nonumber \\
&\left.{\sum}^{-1}
  ({\bf x}_{t_{uj} = t}-\bm{\mu}) \right)
\end{align} 

where $\bm{\mu}$ is an $n$-element vector with all entries equal to $\mu_P$
(the posterior mean), and $\sum$ is the covariance matrix for which all diagonal
elements are equal to $\sigma^2 + \sigma_P^2$ and all off diagonal
elements are equal to $\sigma_P^2$. 

We can also analytically obtain the determinant of the
covariance matrix $\sum: \left|\sum\right|= \sigma^{2n} + n
\sigma^{2(n-1)} \sigma_P^2$, where $n$ is the dimensionality of the
matrix or the number of customers sitting at the table whose dish we
are sampling for.  
\section{Experiments}
\label{sec:experiments}
The proposed iM$^3$F model was evaluated on the MovieLens and DBLP datasets.  On the
MovieLens set, we
broadly divided our results into internal comparisons, meaning
hyperparameter variation experiments, and external comparisons, meaning the
improvements vis-$\grave{\textrm{a}}$-vis M$^3$F.  For the DBLP
dataset, we present final numbers in comparison to M$^3$F.  While
previous work evaluated their approaches on the Netflix Prize dataset,
we are unable to do so due to the unavailability of the data.  All experiments
were run on a quad-core 2.5 GHz Linux machine with 8GB RAM, and the
implementation was done in MATLAB and C.  
\subsection{Datasets}
The MovieLens\footnote{http://www.grouplens.org} dataset is a movie
rating set, similar to those used in the Netflix prize. We used two versions of the datasets (ml100k
and ml1m) that contain
100,000 and 1 million ratings respectively. Table \ref{tab:stat} shows
some summary statistics.
\begin{table}[t]
 \begin{center}
    \begin{tabular}{ccccc}
      {\bf Name}  &{\bf Ratings} & {\bf Users} & {\bf Movies} & {\bf Sparsity} \\ \hline \\
      ml100k & 100,000 & 943 & 1682 & 6.3\% \\
      ml1m & 1,000,000 & 6040 & 3952 & 4.2\% 
    \end{tabular}
  \end{center}
  \caption{MovieLens Dataset Statistics.  Sparsity refers to the percentage of
    non-zero elements in the matrix.}
  \label{tab:stat}
\end{table}

In both datasets, each user has rated at least 20
movies, with each rating being an integer from 1 to 5. When comparing
the performance of different models, we measured the Root Mean Square
Error (RMSE) on a held-out test set that is generated using a
leave-one-out strategy: for each user, one random rating is chosen out
of her movie ratings and put into the test set; the rest remain in the
training set.  We performed this random split twice, and all results
for the MovieLens experiments are averaged over the two splits. 

We have also extracted a publication dataset from the latest DBLP XML data\footnote{http://dblp.uni-trier.de/xml/} as of January 6, 2013.  The
extracted co-authorship dataset can be made publicly available for comparison
purposes.  The dataset consists of around 4.5 million ratings; each
rating is between two authors, and is simply a count of the number of
papers that they have co-authored.  For the purposes of our experiments, we selected
the first 2 million ratings between 1 and 10, which consisted of
644,256 authors, and split the dataset
into training and test components such that every author is
represented at least once in the test set.  In this
dataset, the interpretation of the topics is akin to clusterings of
researchers based on a specific sub-field, or geographical location.  
\subsection{Internal Comparisons}
We use the ml100k dataset with 100 sampling iterations
to look at how varying the hyperparameters $\beta$ and $\gamma$ affect
the final number of topics estimated from the data.  The larger ml1m
dataset with 500 sampling iterations was used to evaluate RMSE for both the hyperparameter
variation experiments and the final comparison with M$^3$F.  When comparing
against M$^3$F, we use the hyperparameter settings that yield the best
performance as in the original paper \cite{Mackey2010}: 
\begin{itemize}
  \item Number of user topics $K^U$: 2
    \item Number of item topics $K^M$: 1
      \item Latent factor dimensionality $D$: 40
\end{itemize}
As we do not alter the latent factor terms, we maintain $D=40$ for the
iM$^3$F model. 
\begin{figure}[t!]
  \begin{subfigure}[t]{0.5\textwidth}
    \centering
    \includegraphics[width=0.99\linewidth]{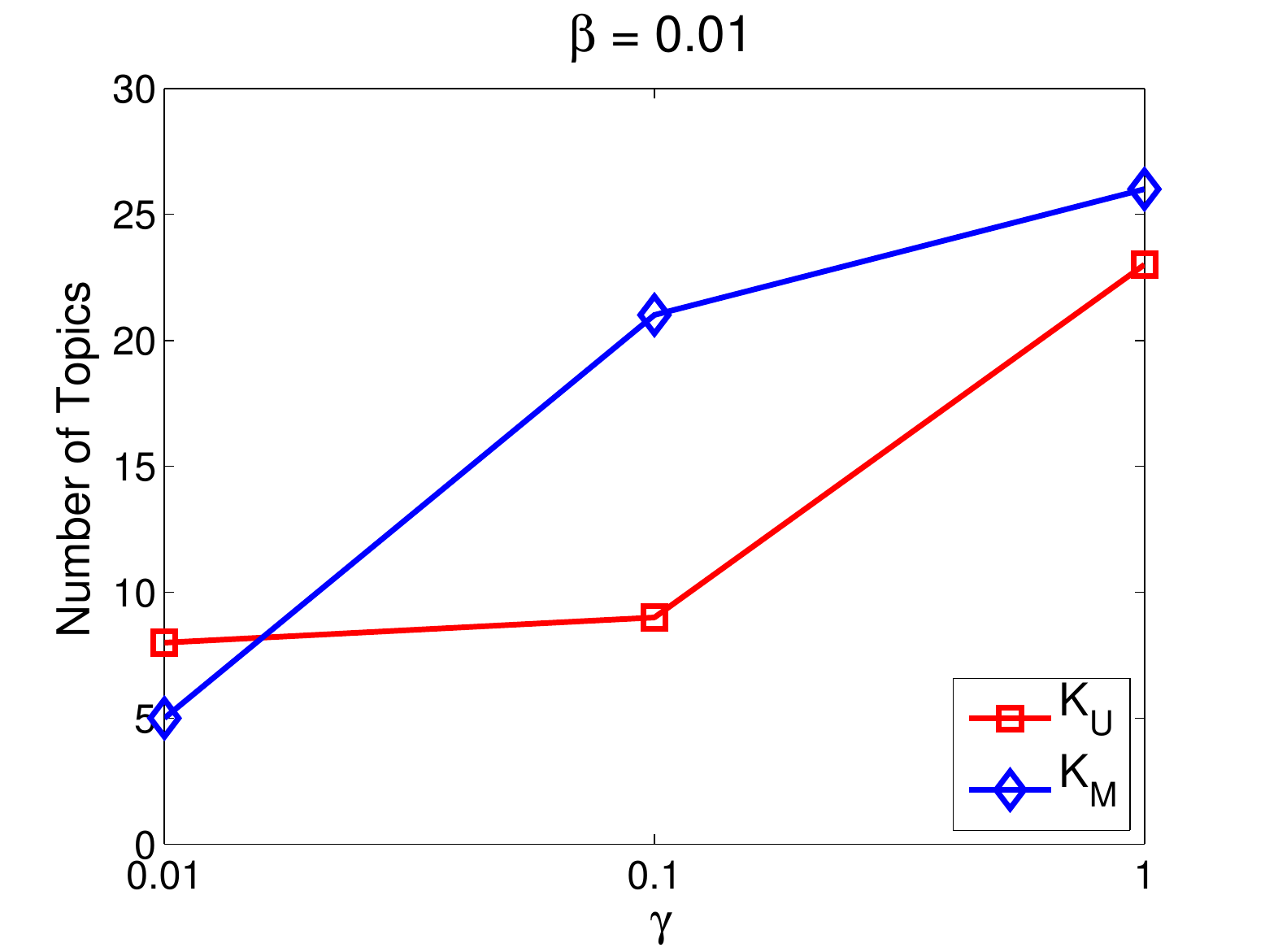} 
    \caption{Final topic number of CRF model, $\beta=0.01$}
    \label{fig:beta0.01}
  \end{subfigure}%
  \\
  \begin{subfigure}[t]{0.5\textwidth}
    \centering
    \includegraphics[width=0.99\linewidth]{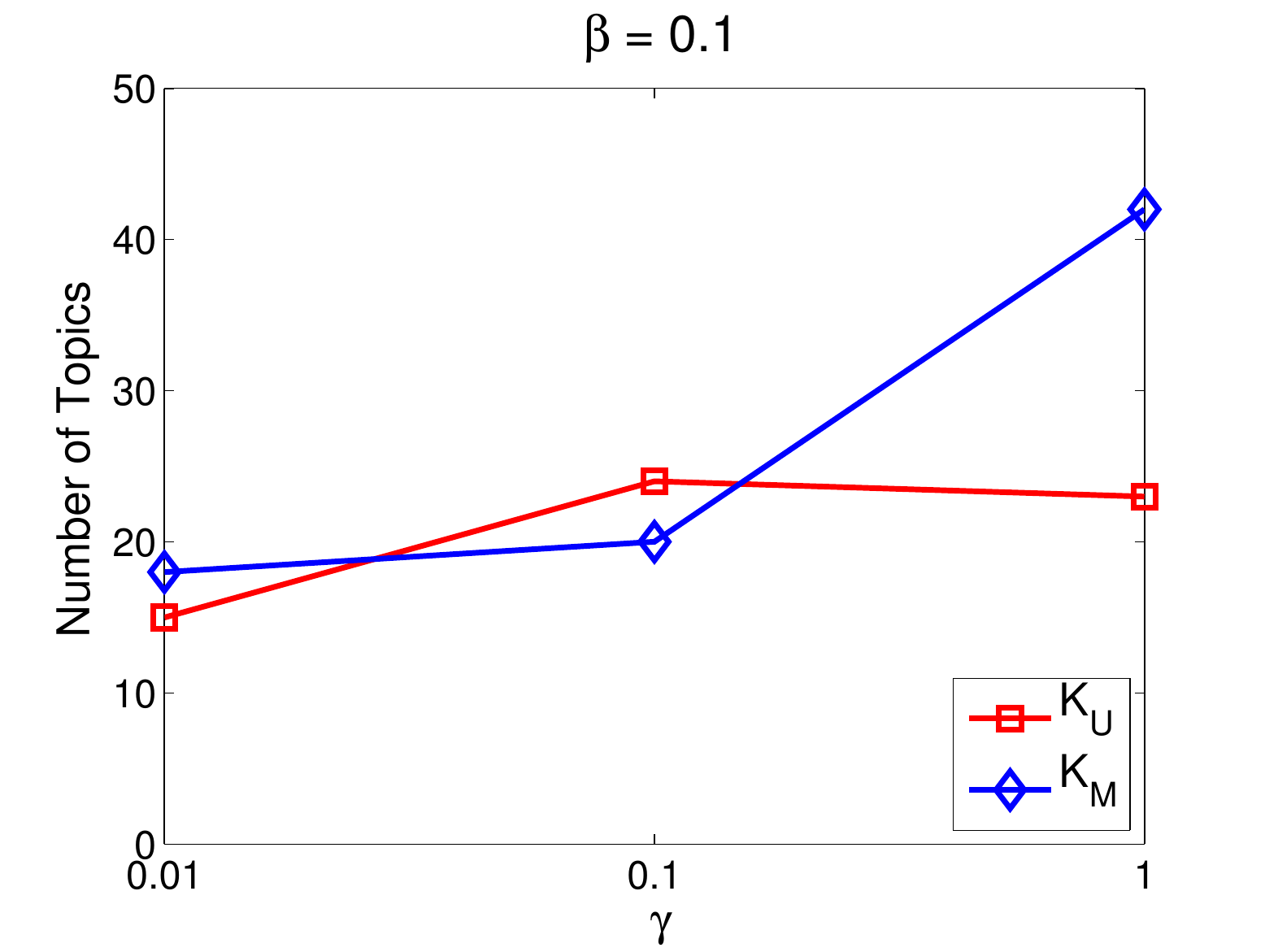} 
    \caption{Final topic number of CRF model, $\beta=0.1$}
    \label{fig:beta0.1}
  \end{subfigure}%
  \\
 \begin{subfigure}[t]{0.5\textwidth}
    \centering
    \includegraphics[width=0.99\linewidth]{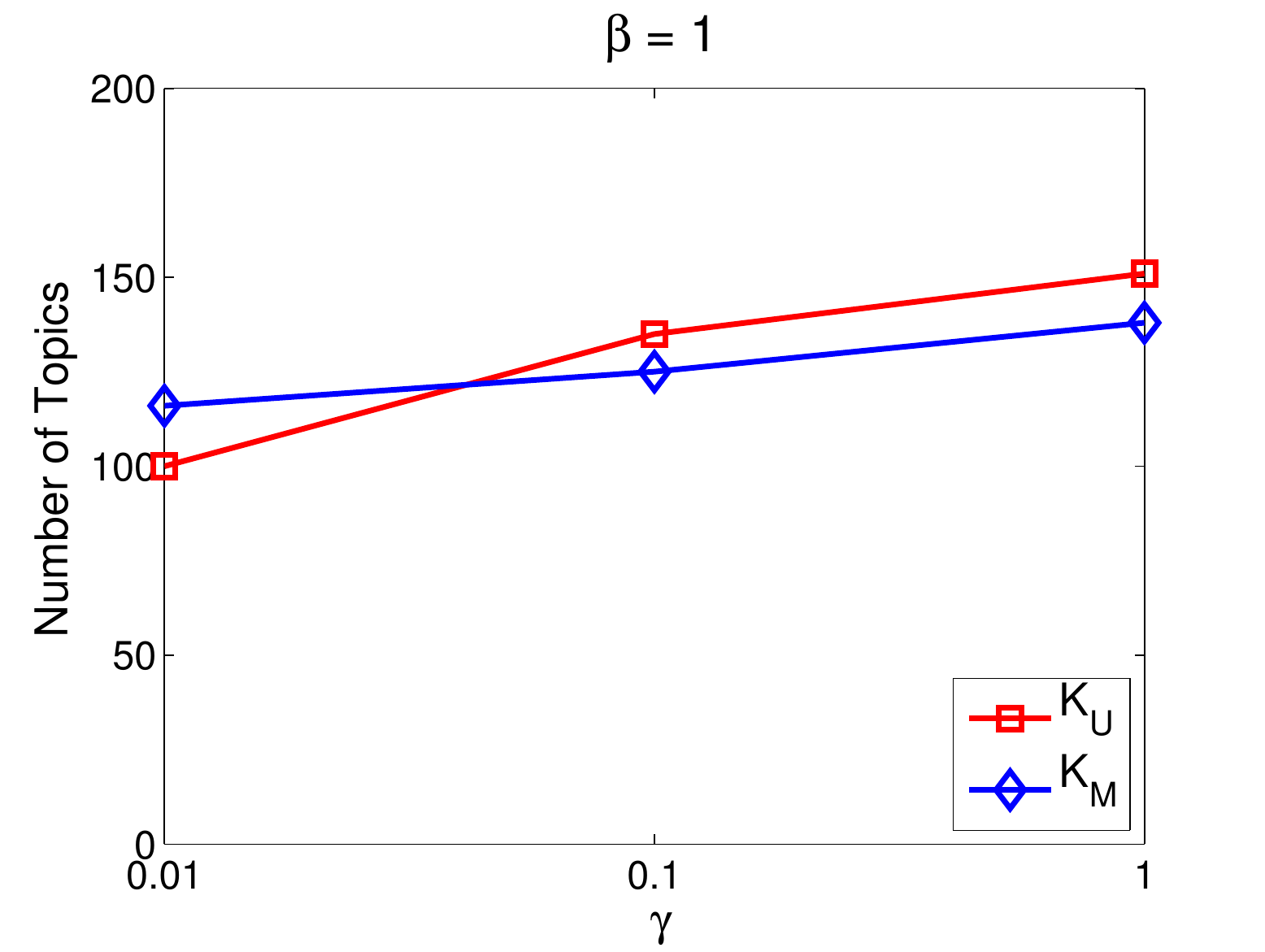} 
    \caption{Final topic number of CRF model, $\beta=1$}
    \label{fig:beta1}
  \end{subfigure}%
  \caption{Impact of hyperparameters on CRF model}
  \label{fig:hyperparameter}
\end{figure}
The impact of our CRF concentration parameters $\beta$ and $\gamma$,
which correspond to the CRPs of the dish and table processes respectively,
are presented in Figure \ref{fig:hyperparameter}.  In Figures
\ref{fig:beta0.01} to \ref{fig:beta1}, we show the final topic numbers
after running 100 iterations of the Gibbs sampler for the
CRF model as we varied $\gamma$ and $\beta$.  The overall trend is as
expected, with higher $\beta$ and $\gamma$ values resulting in more
topics.  
\begin{figure}[h]
  \begin{subfigure}[b]{0.5\textwidth}
    \centering
    \includegraphics[width=0.99\linewidth]{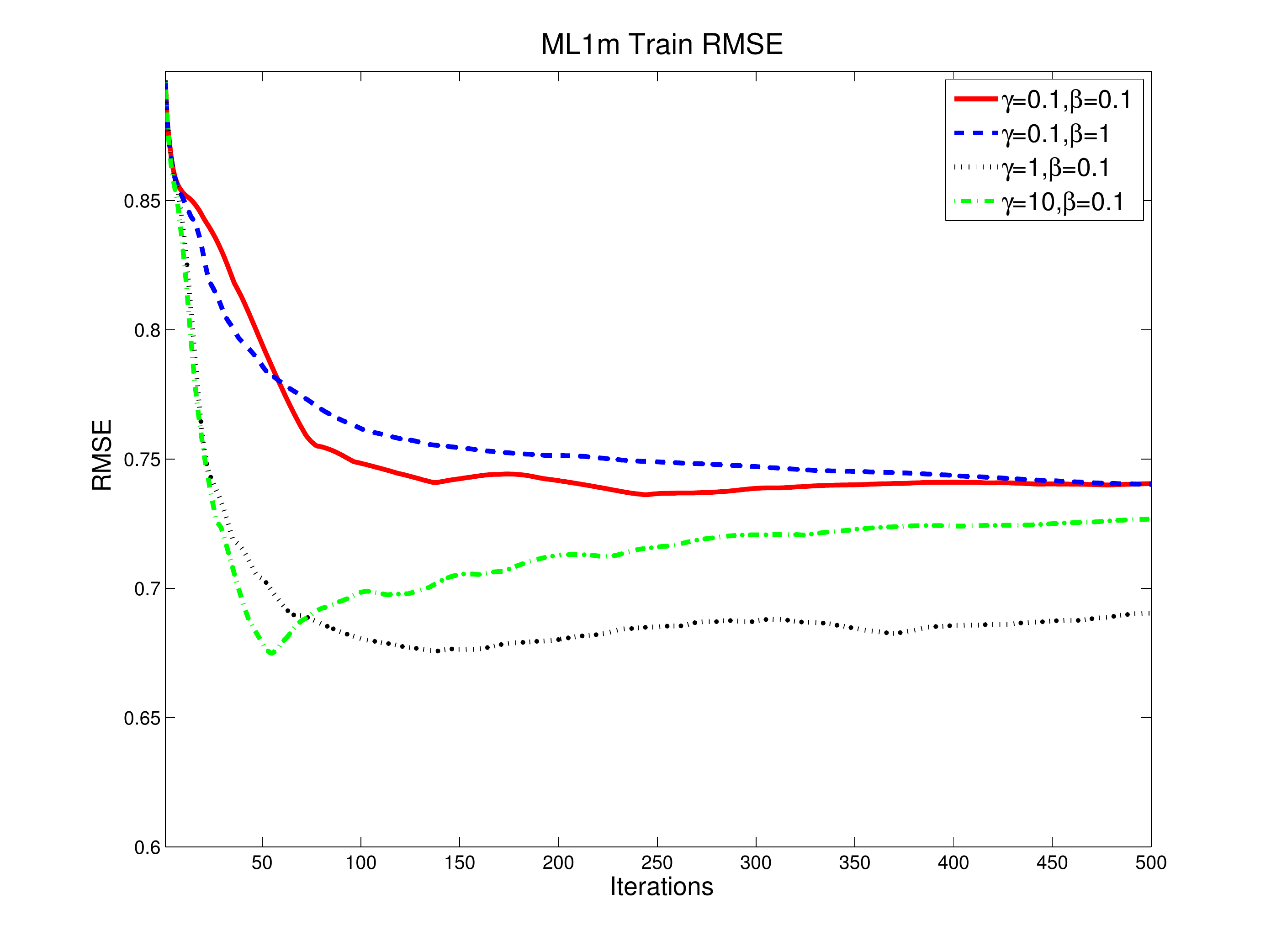} 
    \caption{ML1M, Train Set RMSE}
  \end{subfigure}%
  \\
  \begin{subfigure}[b]{0.5\textwidth}
    \centering
    \includegraphics[width=0.99\linewidth]{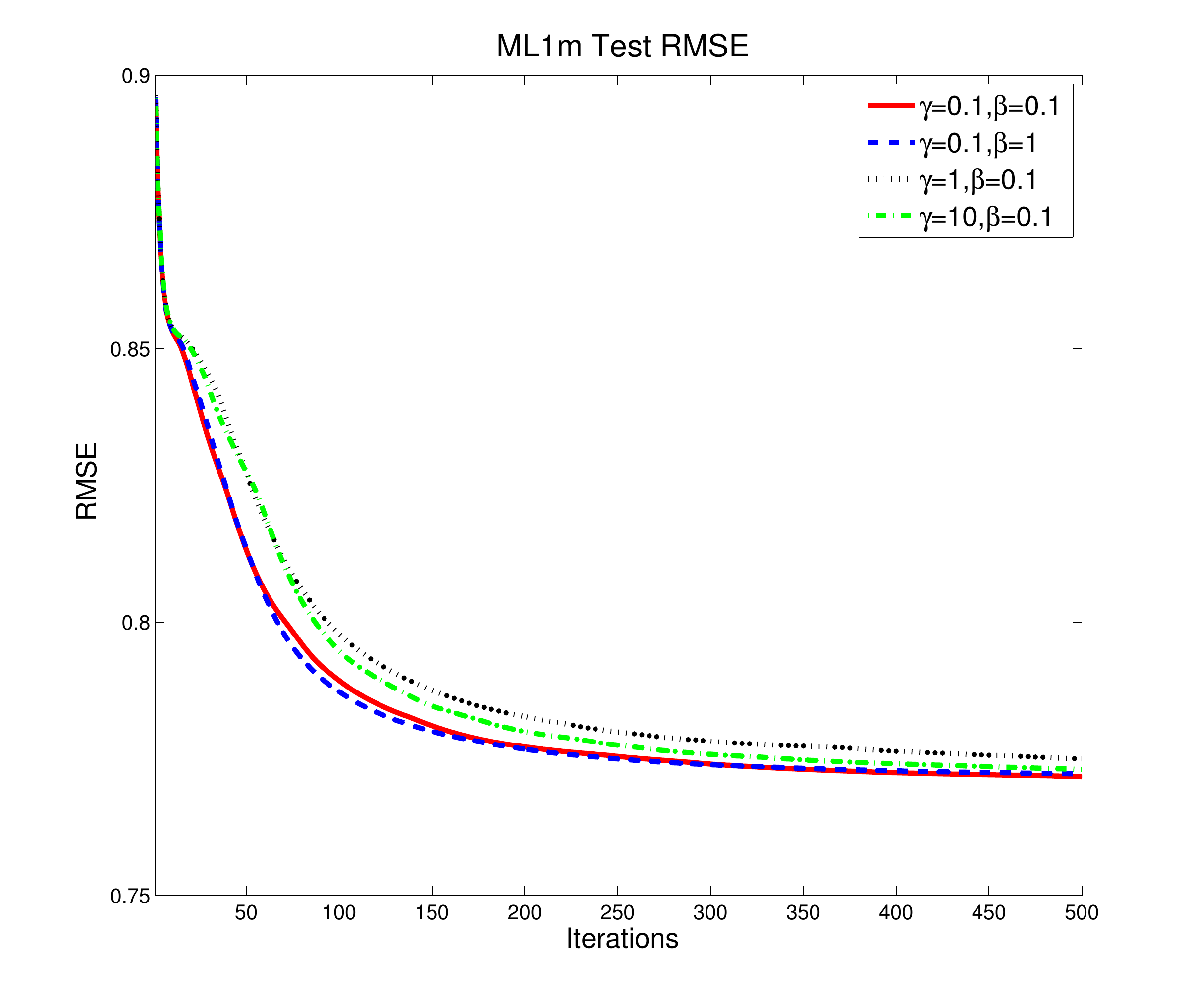} 
    \caption{ML1M, Test Set RMSE}
   \end{subfigure}
  \caption{RMSE comparison of various iM$^3$F models.}
  \label{fig:ml1mcrftraintestrmse}
\end{figure}
Figure \ref{fig:ml1mcrftraintestrmse} presents the train and
test set RMSE on ml1m for certain settings of the concentration parameters $\beta$ and
$\gamma$.  The lowest training RMSE was obtained when $\gamma = 1,
\beta=0.1$ (0.6904), with $K^U = 94.5$ and $K^M = 100$, which we refer to as `TrainBest' in subsequent
exposition.  The lowest test RMSE was when $\gamma =
0.1, \beta = 0.1$ (0.7718), i.e., 'TestBest', with $K^U = 64$, and
$K^M = 77.5$.\footnote{fractional values are obtained due to averaging
  between the two ML1M splits}
\subsection{iM$^3$F and M$^3$F}
\begin{figure}[h]
    \centering
    \includegraphics[width=0.99\linewidth]{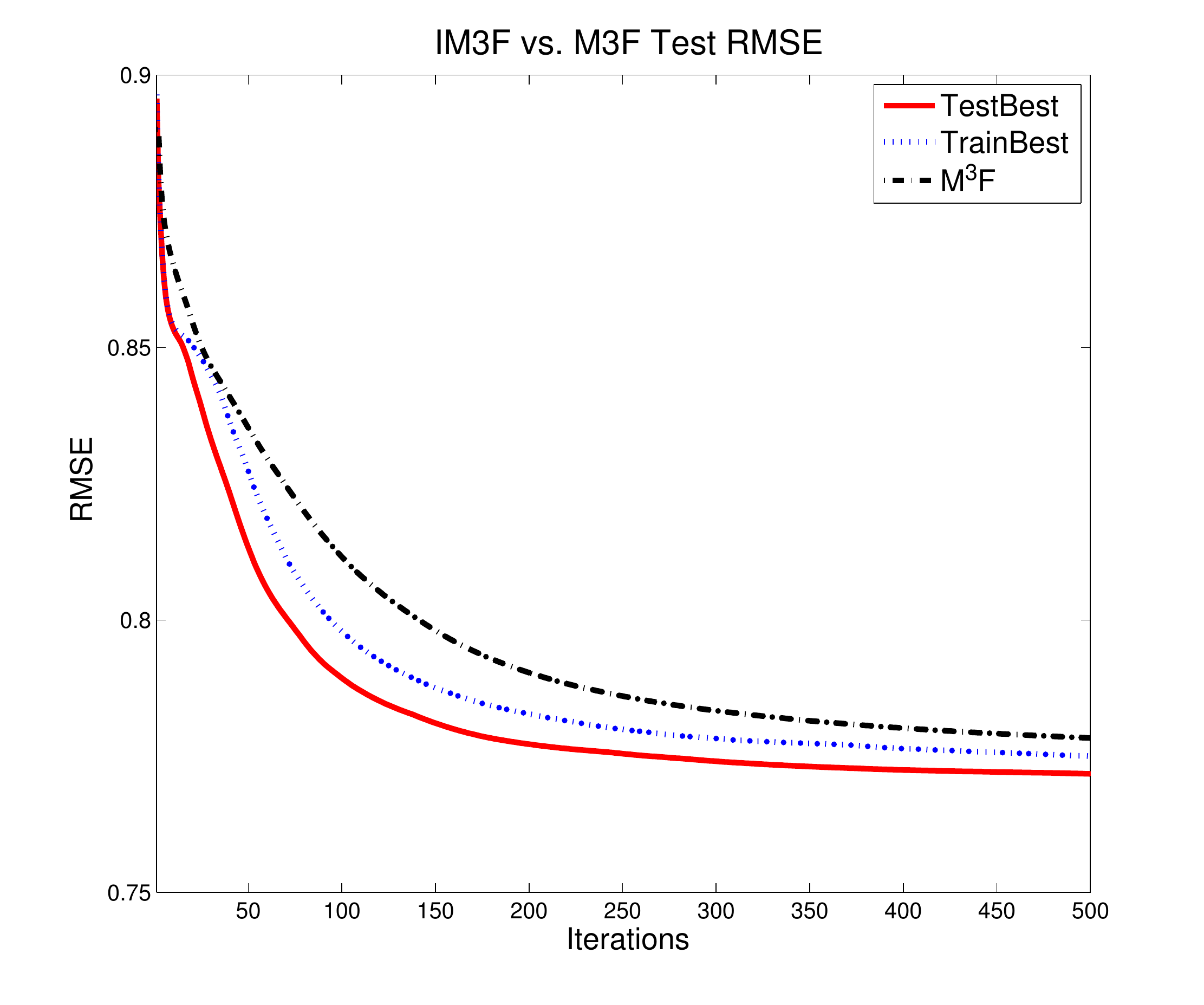} 
 \caption{Test RMSE Comparison of iM3F and M3F model on the  ml1m dataset.}
  \label{fig:external}
\end{figure}

We then compared our CRF iM$^3$F model with the M$^3$F
implementation of \cite{Mackey2010}.  For the ml1m data, the M$^3$F model also
shows a better fit to the data, and achieves a much lower train RMSE
of 0.60. However, in the test set comparison in
Figure \ref{fig:external}, both of our CRF models outperformed M$^3$F after 500 rounds of Gibbs
sampling. The final test RMSE for `TestBest' (our
best result) was {\bf 0.7718}, while for the M$^3$F model it was {\bf
  0.7783}\footnote{note that this is different than the result on the
  same dataset published in previous work, however, we were unable to
  replicate the previous number despite running the original
  implementation directly.}, the difference being statistically significant according to
the paired $t$-test.  To put this improvement in context, the test set gain of
0.0065 is more than twice the
improvement that M$^3$F achieved over its baseline, BPMF, the
difference between M$^3$F and BPMF being the addition of the MMSB to
model the contextual bias.  `TrainBest'
achieved a test RMSE of 0.7750, also an improvement greater than the M$^3$F-BPMF improvement. 

\begin{table}[h!]

  \begin{center}
    \begin{tabular}{p{0.2\linewidth}cccc}
      {\bf Model}  &{\bf Train RMSE} & {\bf Test RMSE} & {\bf $K^U$} &
      {\bf $K^M$} \\ \hline \\
      M$^3$F & 0.935  & 1.149  & 2 & 1\\
      iM$^3$F, $\gamma=0.1,\beta=0.1$ & 0.600 & 1.075 & 6 & 5 \\
      iM$^3$F, $\gamma=0.1,\beta=1$ & 0.544 & 1.049 & 23 & 27
    \end{tabular}
  \end{center}
  \caption{Final train and test set RMSE results on the DBLP dataset.
  Both iM$^3$F models obtain better train and test set RMSE.}
 \label{tab:dblpresults}
\end{table}

On the DBLP dataset, for M$^3$F we carried out an extensive three-way
grid search, varying the number of latent factors (with the values
$\{5, 10, 15, 20, 25, 30\}$), and user and item
topics (with the values $\{1, 2, 4, 8, 16, 32\}$), and found the best configuration to be $D=25, K^U = 2,
K^M=1$.  The very fact that we had to implement an extensive grid
search for these parameters already indicates the potential advantages
of the nonparametric approach.  For the iM$^3$F model, guided by previous results we set
$\gamma=0.1$ and tried $\beta=0.1$ and 1.  The final results are
presented in Table \ref{tab:dblpresults}, along with the final number of
user and item topics.  It is interesting to note that the number of
user and item topics that achieved the best test RMSE in iM$^3$F is
quite a bit higher than the optimal parameters for M$^3$F.  

The performance gain achievable through iM$^3$F comes at a
cost, in that Gibbs sampling with nonparametric models
is computationally more expensive than with parametric ones, as one
always has to consider growing the number of clusters in light of more
data.  
However, we note that recent efforts that develop exact,
distributed MCMC procedures for Dirichlet process-based models
\cite{Williamson2013} can easily be applied in our scenario to
accelerate Gibbs sampling, which is a simple solution for today's
multicore computers.  We emphasize however, that we have not
added additional parameters to the model; rather, $K^U$ and $K^M$ have
been replaced with a pair of hyperparameters, $\beta$ and $\gamma$,
providing the model more flexibility to learn.

\section{Future Work \& Conclusion}
\begin{figure}[h] 
  \begin{center}
    \includegraphics[scale= 0.35]{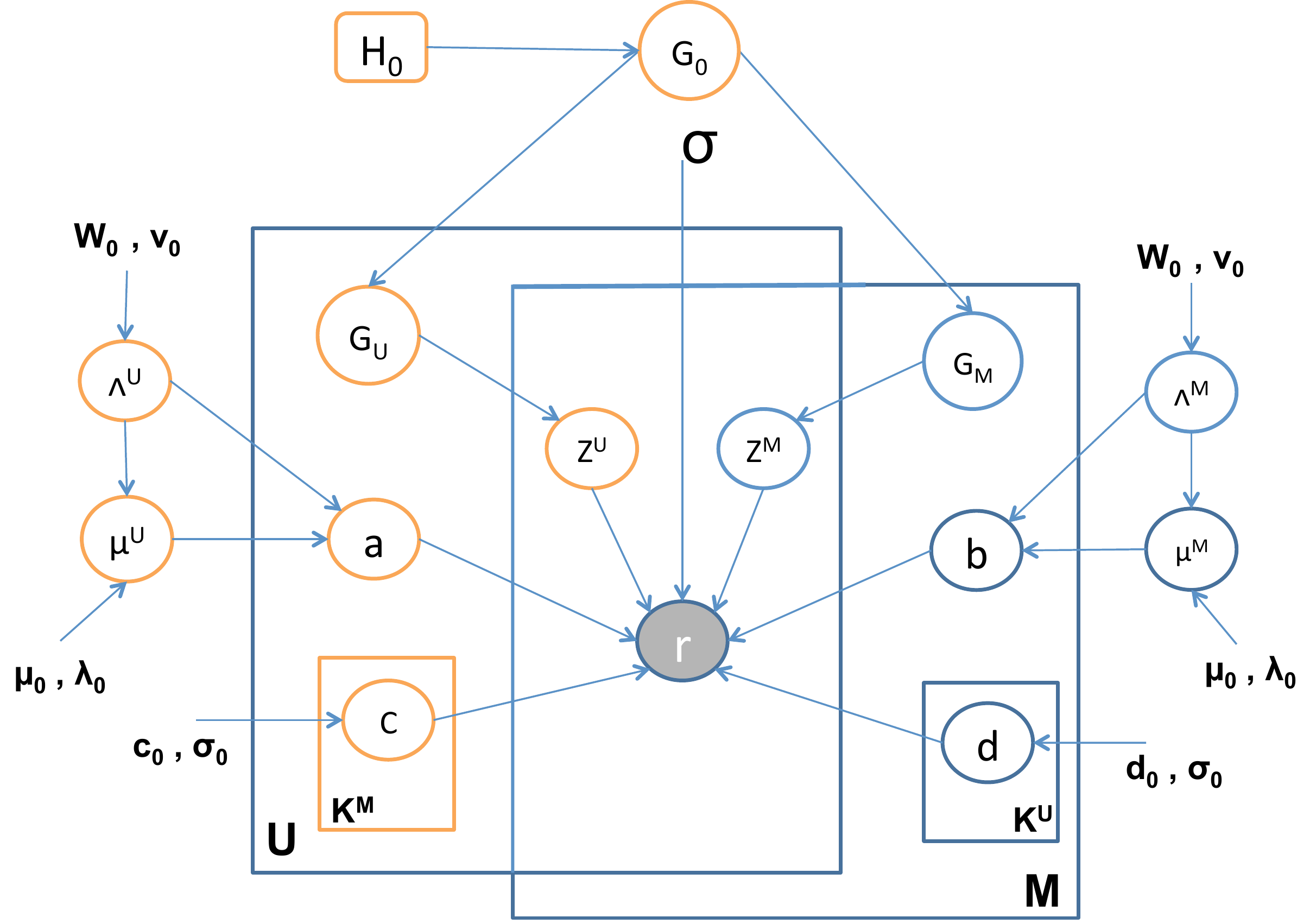}
  \end{center}
  \caption{The iM$^3$F model modified to handle reciprocity.  We have a
    single user topic HDP prior shared on both sides of the rating.}
  \label{IMG:M3F_Reciprocal} 
\end{figure}

With the CRF prior, it is also possible to experiment with variants
based on the nature of the data itself.  For example,
\emph{reciprocal} datasets, wherein users rate other users ratings are based on mutual
compatibility, are becoming increasingly more
widespread, e.g., academic publication databases like our DBLP
dataset or the Microsoft
Academic Scholar database, or online dating websites like eHarmony or
OkCupid.  In the future, we would like to model the co-authorship dataset
released (and other similar datasets) in more direct `relational'
manner. The bidirectional ratings need to be combined somehow to
incorporate the principle of reciprocity in these ratings. Handling
such modifications is straightforward in our work, e.g., Figure
\ref{IMG:M3F_Reciprocal}.  

In this paper, we designed and implemented an important addition to the
general field of probabilistic matrix factorization.  We proposed a
nonparametric prior to the mixed membership matrix factorization
model, and in the process derived a Gibbs sampler
using a Chinese restaurant franchise representation.  We then
validated our model by decreasing RMSE on the MovieLens 1M dataset
compared to M$^3$F's RMSE, a much more significant improvement than
what had been done previously in the literature.  We also extracted a
co-authorship database from DBLP, and showed RMSE improvements over
M$^3$F there as well. 

What the results suggest is that when utilizing and
incorporating notions of mixed membership topic modeling into
probabilistic matrix factorization, a nonparametric prior
is a key assumption to make and model, let alone the convenience factor
of not having to perform a grid search to find parameter values.

\section*{Acknowledgment}
We would like to thank Lester Mackey for assistance in modifying his
original codebase to the nonparametric version, Eric P. Xing for his
comments, and the anonymous reviewers for their feedback.  
\newpage
\bibliographystyle{IEEEtran}
\bibliography{refs} 

\end{document}